\documentclass{article}

\usepackage{PRIMEarxiv}
\usepackage[utf8]{inputenc} 
\usepackage[T1]{fontenc}    
\usepackage{hyperref}       
\usepackage{url}            
\usepackage{booktabs}       
\usepackage{amsfonts}       
\usepackage{nicefrac}       
\usepackage{microtype}      
\usepackage{lipsum}
\usepackage{fancyhdr}       
\usepackage{graphicx}       
\graphicspath{{media/}}     
\usepackage{amsmath}

\usepackage[CaptionAfterwards]{fltpage}

\title{Confound-leakage: Confound Removal in Machine Learning Leads to Leakage}

\graphicspath{ {./figures/} }
\usepackage{array}
\usepackage{makecell}

\usepackage{longtable}
\usepackage{capt-of}

\begin{document}

\author{
\\
Sami Hamdan\textsuperscript{1,2},
Bradley C. Love\textsuperscript{3,4,5},
Georg G. von Polier\textsuperscript{1,6,7},
Susanne Weis\textsuperscript{1,2}, \\
Holger Schwender\textsuperscript{8},
Simon B. Eickhoff\textsuperscript{1,2},
Kaustubh R. Patil\textsuperscript{1,2,*}
\\
\bf{1} Institute of Neuroscience and Medicine, \\ \bf{} Brain and Behaviour (INM-7), Forschungszentrum Jülich, Düsseldorf, Germany
\\
\bf{2} Institute of Systems Neuroscience, \\ \bf{} Heinrich-Heine University Düsseldorf, Düsseldorf, Germany
\\ 
\bf{3}  Department of Experimental Psychology, \\ \bf{} University College London, London, UK
\\
\bf{4} The Alan Turing Institute, \\ \bf{} London, UK
\\
\bf{5} European Lab for Learning \& Intelligent Systems (ELLIS)
\\
\bf{6} Department  of  Child  and  Adolescent  Psychiatry, \\ \bf{} Psychosomatics  and Psychotherapy,  University  Hospital  Frankfurt,  Frankfurt,  Germany
\\
\bf{7} Department  of  Child  and  Adolescent  Psychiatry, \\  \bf{} Psychosomatics  and  Psychotherapy, RWTH  Aachen  University, Aachen, Germany
\\
\bf{8} Institute of Mathematics, \\ \bf{} Heinrich-Heine University Düsseldorf, Düsseldorf, Germany
\\
\bigskip
* k.patil@fz-juelich.de
}

\maketitle

\begin{abstract}
Machine learning (ML) approaches to data analysis are now widely adopted in many fields including epidemiology and medicine. 
For meaningful application of these approaches, confounds must first be removed as is commonly done by featurewise removal of their variance by linear regression before applying ML. 
Here, we show this common approach to confound removal biases ML models, leading to misleading results.
Specifically, null or moderate effects are erroneously turned in to near-perfect prediction due to leaked information after deconfounding when nonlinear ML approaches are subsequently applied.
We identify and evaluate possible mechanisms for such confound-leakage and provide practical guidance to mitigate its negative impact. 
We demonstrate the danger of confound-leakage in a real-world clinical application where the accuracy of predicting attention deficit hyperactivity disorder (ADHD) is overestimated when using depression as a confound. 
Our results have wide-reaching implications for implementation and deployment of ML workflows and beg caution against naïve use of standard confound removal approaches.
\end{abstract}


\newpage{}
\section{Introduction}\label{sec1}
Machine learning (ML) approaches have revolutionized data analysis by providing powerful tools, especially nonlinear models, that can model complex feature-target relationships. 
However, the very power these nonlinear models bring to data analysis also lead to new challenges.
Specifically, as we will detail, when a standard confound removal approach is paired with nonlinear models, new and surprising issues arise as the unintended is discovered and misinterpreted as a true effect. 

Imagine building a diagnostic classifier for attention deficit hyperactivity disorder (ADHD) based on speech patterns. 
This will be a useful clinical tool aiding objective diagnosis \cite{Polier2021}. 
However, like most disorders, ADHD has comorbidity, for instance with depression. 
Ideally, an ADHD diagnostic classifier should only rely upon characteristics of ADHD and ignore that of depression. 
This is an example of confounding, where it is desirable that the confound depression is disregarded by the classifier. 
Another example of confounding is the effect of ageing and neurodegenerative diseases on the brain. 
In a study to build a neuroimaging-based diagnostic classifier, the non-pathological ageing signal is confounding \cite{Dukart2011}. 
Confounding is ubiquitous and further examples include batch effects in genomics \cite{Jo2020, Johnson2017, Whalen2022}, scanner effects in neuroimaging \cite{Pomponio2020}, patient and process information in radiographs \cite{Badgeley2019}, and group differences like naturally different brain sizes in investigation of brain-size-independent sex differences \cite{Luders2014, Wiersch2022}. 
Ignoring confounding effects in an ML application can render predictions untrustworthy and insights questionable \cite{Mehrabi2021} as this information can be exploited by learning algorithms \cite{MacKinnon2000} leading to spurious feature-target relationships \cite{Pourhoseingholi2012}, e.g., classification based on depression instead of ADHD or age instead of neuronal pathology. 
The benefits of big data in ML applications are obvious, especially when modeling weak relationships, but big data also leads to an increased risk of inducing confounded models \cite{Dukart2011, Deng2010, AlfaroAlmagro2021, Wiersch2022}.
Confounding, thus, is a crucial concern and if not properly treated can threaten real-world applicability of ML. 

When confounding masks the true feature-target relationship, its removal can clean the signal of interest leading to higher generalizability, e.g. removal of batch effects in genomics \cite{Whalen2022}. 
On the other hand, when confounding introduces artefactual relationships the same procedure can reduce prediction accuracy \cite{Rao2017, Chyzhyk2022}. 
In either case, removing or adjusting for confounding effects is crucial for obtaining unbiased results, as otherwise a ML model might mostly rely on confounds, rendering signals of interest redundant.
Two methods for treating confounding are commonly employed in data analysis. 
Data can be stratified based on the confounding variables, but it may introduce confounding information \cite{Greenland2003}, falsely increase test-set performance by removing harder to classify data points \cite{Snoek2019}, and can result in excessive data loss. 
As confounds share variation -usually presumed linear variance- with both the target and the features, another common method is confound regression (CR) which removes the confounding variance, also called confounded signal, from each feature separately using a linear regression model \cite{Snoek2019, Dukart2011}. 
The resulting residualized features are considered confound-free and are used for subsequent analysis. 
CR has become the default method to counter confounding in observational studies, including in ML applications \cite{Snoek2019, Dinga2020, AlfaroAlmagro2021}.
Typically, a two-step CR-ML workflow is constructed while avoiding risks associated with typical data-leakage by applying CR in a cross-validation-consistent manner \cite{Snoek2019, More2021}.
It is important to note that, we use a practitioner-oriented operational definition of confounds as a set of variables suspected to share an unwanted effect with both the features and target, which does not imply causality as in more formal definitions \cite{VanderWeele2013}.

A CR-ML workflow typically attenuates prediction performance as it removes variance from the features that is informative of the target. 
If an increase in performance is observed after CR, it can be explained by either (1) \textit{information-reveal}: CR reveals information that was masked by confounding or (2) \textit{confound-leakage}: leakage of confounding information into the features.
In the case of information-reveal, CR could suppress linear confounding or noise in turn enhancing the underlying (non-)linear signal and making learning easier for a suitable ML algorithm \cite{MacKinnon2000}. 
This would be a positive effect similar to removing simple shortcuts in the data \cite{Dagaev2021, Geirhos2020}.
If this is the case then the resulting CR-ML workflow would be a valuable for modelling non-linear relationships.
Alternatively, as CR is a univariate operation applied to each feature, multivariate confounding (across features) could be revealed, which could help prediction albeit undesirably. 
On the other hand, confound-leakage would be an even more worrisome outcome as it would leak confounding information into the features instead of removing it.
Confound-leakage would be detrimental to the validity and interpretability of the ensuing CR-ML workflow and in some cases could lead to dangerous outcomes. 
CR has been reported to induce biases into statistical workflows, albeit not incorporating ML, leading to incorrectly inflated group differences inference in combined batch effects removal and group difference analysis \cite{Nygaard2016}.
Although a recent study has speculated on the pitfall of confound-leakage in ML workflows \cite{Chyzhyk2022}, it has not yet been systematically shown, analyzed nor explained.

To disentangle the two possible explanations of performance increase after CR, we systematically analyzed the two-step CR-ML workflow. 
For analysis purposes and to gain detailed knowledge, we propose a framework that uses the target as a confound (TaCo), in which we use a single confound that is the target.
As a confound needs to share variation with both the target and the feature, any possible confound must share all confounded signal with the target.
Hence, the target can be seen as a “superconfound” subsuming all possible confounding effects. 
Although it is unlikely to encounter a confound equal to the target in real applications, TaCo provides a framework for systematic evaluation.
It should be noted that real confounds will fall on the continuum from weak (low confounded signal) to strong (TaCo) depending on their degree of similarity with the target. 
Indeed, as we show, the TaCo framework reveals strong effects where the prediction accuracy is boosted from moderate to perfect as well as weaker effects for confounds weakly correlated with the target.

To this end, we performed extensive empirical analyses on several benchmark datasets providing strong evidence for confound-leakage.
First, we showcase confound-leakage in a walk-through analyses. 
Then using the TaCo framework we systematically answer whether the improvement in prediction performance after CR is due to leakage. 
For this, we used benchmark datasets as well as several conceptually simple simulations covering both classification and regression problems. 
Finally, with a clinically-relevant task of ADHD diagnosis using speech-related features with depression as a confound, we demonstrate misleading impact of confound-leakage.

\section{Results}\label{sec2}
\subsection{Walk-through analysis
}\label{sec:walk}
The goal of this section is to introduce readers to our analysis approach with intuitive examples.
We show one exemplary case of TaCo removal for a binary classification task and a CR scenario with a weaker confound in a regression task.
In both cases, we randomly split the data into $70\%$ train and $30\%$ test parts.
The CR and prediction models were learned on the training data and the results are reported on the test split.
We will show that, confound-leakage can be concluded if performance using shuffled features after CR ($\tilde{X}_{CR}$) and more importantly confound-predicted-features ($\hat{X}$) is higher than the baseline performance using original features ($X$).

\subsubsection{TaCo removal for binary classification}\label{sec:walk_cr_binary}

We analyzed the "bank investment" data to predict whether a customer will subscribe to term deposit given their financial and socio-economic information.
We used a decision tree (DT) with limited maximum depth of two for visualization ease.
This example is meant to demonstrate key aspects of our proposed analyses (Fig. \ref{fig:walk}). 

TaCo removal showed a much higher area under the curve for the receiver operating characteristic curve (AUCROC)  of 0.98\unskip compared to the baseline AUCROC of 0.75\unskip without CR.
Still, the TaCo removed features were highly similar to the original features (median Pearson's correlation: 0.99\unskip, Fig. \ref{fig:walk} a-b). 
The two ensuing DTs were, however, completely different and relied on different features. 
Notably, these drastic differences were induced by minute feature alterations after CR that are hardly detectable by humans but are effectively captured by DT (Fig. \ref{fig:walk} c-d).
Such performance increase can be either due to revealed information or confound-leakage.
Therefore, we sought to gain evidence to distinguish between these two scenarios using two complementary measurements: 
1) destroying the relationship between features and target, and 
2) use of confound-predicted features.

To destroy feature-target relation we shuffled each feature before CR ($\tilde{X}$) to create $\tilde{X}_{CR}$ and repeated the analysis. 
As there should be no predictive information in the shuffled features, the only explanation for above chance-level performance is CR leaking information into the confound-removed features $X\textsubscript{CR}$, i.e. confound-leakage. We observed chance-level performance without CR ($\text{AUCROC}=0.52\unskip$) for the shuffled features. 
However, a performance increase after TaCo removal was observed ($\text{AUCROC}=0.98\unskip$). 
This analysis shows that performance increase after TaCo removal with shuffled features indicate the possibility of confound-leakage.
Nevertheless, it can be argued that confound-leakage on the shuffled features, does not necessarily imply leakage for the non-shuffled features.
Therefore, we used confound-predicted features $\hat{X}$ to gain direct evidence for confound-leakage using the non-shuffled features. 
In case of information-reveal, an increase in prediction performance after CR is due to removal of noise or weakly informative variance such as linear shortcuts.
This means that the confound-predicted features $\hat{X}$ can only be predicting this weakly/ not informative variance in fact meaning that  $\hat{X}$ can only be at most as predictive as $X$.
In other words, higher accuracy when using $\hat{X}$ than $X$ provides evidence of confound-leakage. In this walk-through example $\hat{X}$ ($\text{AUCROC}=1.0\unskip$) achieved higher prediction score than $X$ ($\text{AUCROC}=0.75\unskip$) providing direct evidence of confound-leakage. 
Together shuffling the features and $\hat{X}$-based prediction clearly demonstrate that the prediction boost is due to confound-leakage rather than information-reveal.

\subsubsection{Confound removal for regression}\label{sec:walk_cr_regression}

As an example of a weaker confound on a regression task, we simulated a binary confound and then sampled a feature from different distributions for each confound value (confound equal to $0$ or $1$). 
Then we added the confound to a normally distributed target ($M=0$ and $SD=0.5$, Fig. \ref{fig:walk} e-f).
This creates a clear confounding situation, where the confound affects both the feature ( $\text{Point-biserial correlation}=0.71\unskip$, $p<0.01$) and the target ($\text{Point-biserial correlation}=0.71\unskip$, $p<0.01$) and thus leads to a spurious relationship between the feature and the target ($\text{Pearson's correlation}= 0.51\unskip$, $p<0.01$).
Following the same procedure as in the previous example, we observed increased performance after CR using a DT with limited depth of two ($R^2$ using $X=0.29\unskip$, $X_{CR}=0.42\unskip$). 
As in this simulated data only a spurious relation (via confound) exists between the feature and target, it is safe to assume that an increased performance after CR is due to confound-leakage.
Still, shuffled features were not sensitive to confound-leakage ($\tilde{X}=0.0\unskip$, $\tilde{X}_{CR}=-0.01\unskip$). 
On the other hand, $\hat{X}$-based predictions clearly indicate confound-leakage ($\hat{X}=0.51\unskip$).
Furthermore, we found a probable mechanism behind this confound-leakage to be the distribution of the features conditioned on the confound (see \ref{sec:walk_cr_regression}).
More precisely, CR shifts the feature values for $\text{confound}=1$ in between most feature values for the $\text{confound}=0$ (Fig. \ref{fig:walk} e).
This leaks the confounding information into the feature instead of removing it (Fig. \ref{fig:walk} f).

\subsection{Analyses of benchmark data
}\label{sec:benchmark}

\subsubsection{TaCo removal increases performance of nonlinear methods}\label{sec:benchmark-TaCo}
Our systematic and CV-consistent analysis comprised comparison between TaCo removal pipelines and no-CR pipelines on 10 UC Irvine (UCI) datasets .
TaCo removal led to a meaningful increase in out-of-sample scoring using all tested non-linear models, RF ($7/10$ datasets), DT ($8/10$) SVM with RBF kernel ($5/10$) and MLP ($7/10$) (Fig. \ref{fig:uci}, Supplementary Fig. S1).
This suggests that confound-leakage is a risk associated with the usage of a CR-ML pipeline with non-linear ML models.
Furthermore, this suggests that the DT-based algorithms (DT and RF) are most susceptible to showing increased performance.

\subsubsection{CR using weaker confounds also increases performance}\label{sec:benchmark-weaker}
As the target is the strongest possible confound, TaCo represents an extreme case. 
To test whether the potential leakage we found with TaCo extends to CR in general, using the UCI datasets we simulated confounds related to the target at different strengths measured by Pearson’s correlation ranging $0.2-0.8$. 
Depending on the dataset, different amounts of correlated confounds led to leakage after CR. 
We observed potential confound-leakage for $5$ of the $10$ datasets with at least one of the confound-target strengths. 
As expected, a higher target-confound correlation led to more leakage, i.e., higher performance after CR (Fig. \ref{fig:uci} C). 

\subsubsection{Increased performance after TaCo removal is due to confound-leakage}\label{sec:increase_due_to_leak}
As described in the walk-through analysis (\ref{sec:walk_cr_binary}), we measure the performance after first shuffling the features and $\hat{X}$  to evaluate whether the increased performance after TaCo removal/CR is due to information reveal or confound-leakage.
After shuffling the features, both pipelines, no-CR and TaCo removal, should perform close to chance-level if the improved performance is due to revealed information. 
Indeed, the no-CR pipeline performed close to the chance level, while  TaCo removal pipeline increased the performance (Fig. \ref{fig:uci} TaCo CR Shuffled). 
As there should be no predictive information in the shuffled features, above chance-level performance could only be obtained if the CR leaks information. 
Thus this result provides strong evidence in-favor of the confound-leakage. 
Inline with these results, $\hat{X}$ was also able to predict the target better than $X$ (Fig. \ref{fig:uci},  Supplementary Fig. S1).

For the simulated weaker confounds these results were less strong, still we found $5/10$ datasets where $X_{CR}$, $9/10$ where $\tilde{X}_{CR}$ performed above chance-level and $3/10$ where $\hat{X}$ had performed better than $X$.

\subsubsection{ Possible mechanisms for confound-leakage
}\label{subsec2.2.5}
As a multitude of mechanisms could lead to confound-leakage, exhaustively identifying all possible mechanisms is out of the scope of this paper. 
Rather we want to highlight two possible mechanisms leading to confound-leakage inspired by the walk-through analyses:
1) Confound-leakage due to continuous features deviating from normal distributions (\ref{sec:walk_cr_regression})
2) Confound-leakage due to unbalanced features of limited precision (\ref{sec:walk_cr_binary}).
Both mechanisms could be summarized under the umbrella of (small) differences of the conditional distributions of features given the confound inside of CV-folds. 

As DT-based models are very popular ML algorithms \cite{Grinsztajn2022} and seem to be most susceptible to the described problems (\ref{sec:benchmark-TaCo}) we will focus on them in our simulations to decrease the complexity of our results.
Furthermore, we will use a DT whenever there is only one features and RF when there are multiple features. 

\subsubsection{Confound-leakage due to deviation from normal distributions }\label{subsec2.3.1}

Consider simulating a standard normal feature not informative of a binary target. 
Then consider adding a smaller distribution around opposing extreme values separately for each class of a binary target (Fig. \ref{fig:sim} a). The resulting feature only differs systematically w.r.t. the classes at the extreme values. 
As CR with a binary confound is equivalent to subtracting the mean for each confounding group from the respective feature, this operation is now biased towards the extreme parts of the feature distribution.
Consequently, $X_{CR}$ exposes confounding information in terms of decrease in the overlap of the feature distributions conditioned on the confound (Fig. \ref{fig:sim} a-b). 
In other words, confounding information leaked via CR in turn increasing the prediction performance (AUROC from $0.51\unskip$ before to  $0.58\unskip$ after TaCo removal).
To show that the increased performance is not only due to better prediction of extreme values, we also tested the same model on a test set without the extreme values. 
The results were in line with previous observations, as the AUROC improved from $0.48\unskip$ before to $0.57\unskip$ after CR.

We also observed higher performance after similar decreased overlap due to TaCo removal in a simplified version of the "house pricing" UCI benchmark dataset (\ref{fig:sim} c-d), providing real world evidence for this phenomena. 

Lastly, we investigated whether such effects could also occur when randomly sampling non-normal distributed features instead of carefully constructing the features conditioned on the confound. To this end, we sampled an increasing number of features ($1$ to $100$) either using a random normal or skewed ($\chi^2$, $df=3$) distribution independent of a normally distributed target.

Using RF, we observed increased performance after TaCo removal with skewed features but not with normally distributed features, e.g. $R^2$ of $M=0.23\unskip$ with $SD=0.06\unskip$ compared to $R^2$ of $M=-0.04\unskip$ with $SD=0.04\unskip$, respectively with $100$ features. 
Importantly, this effect increased with the number of features ( Supplementary Fig. S2). 
These simulations show that skewed features, and by extension potentially other non-normal distributed features, can lead to confound-leakage.

\subsubsection{Confound-leakage due to limited precision features}\label{subsec2.3.2}

A similar effect was observed with binary features, where unbalanced feature distributions conditioned on the confound led to leakage. 
Using simulations first we confirmed that a binary feature perfectly balanced in respect to the TaCo did not lead to confound-leakage (AUCROC of $M=$ 0.5\unskip, $SD=$ 0.0\unskip). 
Then, we repeated similar simulations but now we swapped two randomly selected distinct values of the feature within each CV-fold, preserving the marginal distribution of the feature but slightly changing its distribution conditional on the confound. 
This can be seen as adding a small amount of noise to the feature. 
Still, such a simple manipulation led to drastic leakage after TaCo removal with perfect AUCROC ($M= 1.0\unskip$, $SD=0.0\unskip$), compared to AUCROC without CR ($M= 0.52\unskip$, $SD= 0.0\unskip$). 

To further demonstrate this effect, we analyzed a simple demonstrative classification task using DT and two binary features derived from the UCI "heart dataset" representing the resting electrocardiographic (Restecg) results. 
Without CR the DT had 117 nodes and achieved a moderate AUROC ($M=0.74\unskip$, $SD=0.06\unskip$). 
In stark contrast, after TaCo removal, the DT was extremely simple with only five nodes and achieved near-perfect AUROC ($M=0.99\unskip$, $SD=0.01\unskip$) (Fig. \ref{fig:sim} E). 
Tellingly, this DT was able to make accurate predictions based on numerically minute differences in feature values. 
The reason for this becomes apparent when remembering that CR with a binary confound is equivalent to subtracting the mean of the corresponding confounding group from the respective feature. 
When applied to a binary feature, this results in four distinct values for a residual feature (Fig \ref{fig:sim} E). 
When taken together with the results on the benchmark UCI data (\ref{sec:benchmark-TaCo}), we can see that such minute differences can be exploited by models such as DTs, RFs and MLPs but likely not by linear models.
It is important to note, that leakage through minute differences was not only observed for binary features, but also other features with a limited precision (values containing only integer or with limited fractional parts). 
To demonstrate this, we predicted a random continuous target using either a normally distributed feature or the same feature rounded to the first digit.
The original non-rounded feature performed at chance level both before ($R^2: M=-1.1\unskip$, $SD=0.06\unskip$) and after TaCo removal ($R^2: M=-1.03\unskip$, $SD=0.07\unskip$), while after rounding it lead to an improvement from $M=-0.08\unskip$ ($SD=0.01\unskip$) to $M=0.7\unskip$ ($SD=0.16\unskip$)  after TaCo removal.
Features with limited precision, i.e. with no or rounded fractional part, are common, for instance, age in years, questionnaires in psychology and social sciences, and transcriptomic data. 

\subsection{Confound-leakage poses danger in clinical applications}\label{subsec2.4}
 ADHD is a common psychiatric disorder that is currently diagnosed based on symptomatology but objective computerized diagnosis is desirable \cite{Gualtieri2005}. 
 Ideally a predictive model for diagnosing ADHD should not be biased by co-morbid conditions, e.g. depression \cite{Katzman2017}. 
 To this end, comorbidity can be treated as a confound. 
 However, a confound-leakage affected model, albeit with appealing performance, could lead to misleading diagnosis and treatment. 
 To highlight the danger of confound-leakage on this clinically relevant task, we analyzed a dataset with speech-derived features with the task to distinguish individuals with ADHD from controls. Our version of the dataset is a balanced subsample of the dataset described by Polier et. al. \cite{Polier2021}.

The baseline RF model without CR provided mean AUROC ($M=0.71\unskip$, $SD=0.02\unskip$).
We then removed four confounds commonly considered for this task, age, sex, education level, and depression score (Beck’s depression inventory, BDI), via featurewise CR in a CV-consistent manner. 
This resulted in a much higher AUCROC ($M=0.86\unskip$, $SD=0.02\unskip$).
This model would be very attractive for real-world application if its performance is true--i.e. not impacted by leakage.
However, as we have shown with our analyses confound-leakage can lead to such performance improvement.
If confound-leakage is indeed driving the performance then this model could misclassify individuals as having ADHD because of confounding effects, e.g. their sex or depression, leading to misdiagnosis and wrong therapeutic interventions.
To disentangle the effect of each confound, we looked at the performance after CR for each confound separately. 
Performing CR with BDI led to a high AUCROC  with original features after CR ($M=0.91\unskip$, $SD=0.01\unskip$),   shuffled features ($M=0.84\unskip$, $SD=0.01\unskip$) and $\hat{X}$ ($M=0.84\unskip$, $SD=0.01\unskip$).
This result revealed that BDI is driving the potential leakage, owing to its strong relation to the target (Point-biserial correlation, $ r= \input{paper_val/real_world_relation_BDI_ADHD_r_pointbiserial}\unskip$, $p < 0.01$). 

These analyses clearly demonstrate that real-world applications could suffer from confound-leakage and users should exercise care when implementing and validating a CR-ML workflow. 

\section{Discussion}\label{discussion}
Here, we exposed a hitherto unexplained pitfall in CR-ML workflows that use featurewise linear confound removal--a method popular in epidemiological and clinical applications. 
Specifically, we have shown this method can counter-intuitively introduce confounding, which can be exploited by some non-linear ML algorithms.

We provide evidence of confound-leakage using a range of systematic controlled experiments on real and simulated data comprising both classification and regression tasks.
First, to establish confound-leakage as opposed to information-reveal (of possibly nonlinear information) as the reason behind increased performance after CR, we proposed the TaCo framework, i.e., using the target as “superconfound”. 
This extreme case of confounding allowed us to establish the existence, the extent, and possible mechanisms of confound-leakage.
Specifically, by comparing the without CR baseline performance with CR after feature shuffling ($\tilde{X}_{CR}$) and features as predicted by the confound ($\hat{X}$), this framework can identify confound-leakage as the cause of increased predictive performance. We then extended the same framework to the more realistic scenario of weaker confounds showing that also there confound-leakage can occur. 

To identify risk factors of confound-leakage, we performed several analyses.
First, we demonstrated a mechanism by which confound-leakage can occur: differences of the conditional distributions of 
features given the confound.
In the case of continuous features, non-normal distributions (e.g., skewed distributions) and in the case of discrete features, frequency imbalances can cause leakage, although other mechanisms could exist. 
Additionally, we show that features of limited precision (e.g., age in years and counts) also showed susceptibility due to this mechanism. 
Lastly, our results showed that the risk of confound-leakage increases with the number of features, which is especially problematic in the era of “big data”, where tens of thousands of features are a norm. 

It is important to note that although similar, confound-leakage is not equal to collider-bias.
Colliders are variables causally influenced by both the features and target \cite{Greenland2003}.
Both collider-bias and confound-leakage describe situations where variable adjustment can lead to spurious relationships between features and target. 
However, the collider bias assumes that the removed variable has to be caused by both the features and the target which is not shared by confound-leakage.
One cannot exclude the possibility of collider removal using CR for many of our experiments as our operational definition of confounds does not include any assumption of causality.
Still, we observe confound-leakage through CR for at least one causally defined confound (see walk-through analysis \ref{sec:walk_cr_binary}) and variables showing relationship only with the target.
Such associations are not covered by the causal relationships described by a collider.
In other words, the mechanisms of confound-leakage can lead to leaked information due to any variable related to the target and not only colliders or causal confounds. 

Taken together, our extensive results show that the commonly used data types and settings of non-linear ML pipelines are susceptible to confound-leakage when using featurewise linear CR. 
Therefore, this method should be applied with care, and the ensuing models should be closely inspected, especially in critical decision domains. 
We concretely demonstrated this using an application scenario from precision medicine by building models for diagnosis of ADHD. 
We found that the attempt to control for comorbidity with depression using CR lead to confound-leakage. 
As many disorders often exhibit severe comorbidity, e.g., AHDH and depression as we demonstrated here but also neurodegenerative disorders are strongly confounded by ageing-related factors \cite{WyssCoray2016} as well as comorbidity in mental disorders \cite{Joshi2013, PlanaRipoll2019}, the issue of confound-leakage should be carefully assessed in all such applications. We recommend the following best practices when applying CR together with non-linear ML algorithms:

1) Assess confounding strength: 
Check the confounds' relation to each feature and the target. 
In general, confounds strongly related to the target pose a greater danger of leaking predictive information. 
Here, we used a straightforward approach of measuring the correlations between the confound and target/feature. 
Other methods can be employed, e.g., proposed by Spisak \cite{Spisak2021}. 
Furthermore, measuring how dependent the predictions of a model are on the confound by permutation testing \cite{Epstein2012, Neto2018} or the approach proposed by Dinga et al. \cite{Dinga2020} can be helpful.

2) Compare performance with and without CR: 
If the performance increases after CR, one should investigate the reason behind the increase.

3) Gain evidence against or in favor of the confound-leakage: 
The procedure of shuffling the features followed by CR as we defined in the TaCo framework can provide clues regarding confound-leakage. 
For more direct evidence, the predictive performance of the confound predicted features ($\hat{X}$) can be assessed. 
It is important to note, however, that while this can provide evidence for confound-leakage, we are not aware of a procedure to definitively exclude confound-leakage as an explanation.

4) Carefully choose alternatives: If confound-leakage seems probable then consider alternative confound adjustment methods. 
Stratification \cite{Snoek2019, McNamee2005} is commonly in conventional machine-learning or unlearning of confounding effects \cite{Dinsdale2021} which is common in deep learning and further general approaches that promote fairness \cite{Mehrabi2021, Zhao2020}. 
Note however, that these procedures may also entail pitfalls. 
Hence, we caution researchers to exercise care when applying any confound adjustment protocol and to carefully consider limitations of the modeling approach used.

\subsection{Conclusions and Future Directions}\label{Limitations}
Important societal questions involving health and economic policy can be informed by applying powerful nonlinear ML models to large datasets. To draw appropriate conclusions, confounds must be removed without introducing new issues that cloud the results. 
In the present study, we performed extensive numerical experiments to gather evidence for confound-leakage.
Using feature shuffling and predictions due to confound predicted features as proposed here, investigators can get an initial indication of whether their pipeline and data are susceptible to confound-leakage. We highlighted the conditions most likely to lead to leakage.
Although we made progress on understanding these issues, there is no full-proof method for detecting and eliminating leakage. We hope our results prompt others to push further, perhaps expanding on the standard definition we adopted for confounds by introducing causal analyses. We hope our and allied efforts inform both researchers and practitioners who incorporate ML models into their data analyses. As a starting point, we suggest following the guidelines we provide to mitigate against confound-leakage.

\section{Methods}\label{sec4}

\subsection{Data}\label{sec4.1}
We analyzed several ML benchmark datasets from diverse domains to draw generalizable conclusions. To ensure reproducibility, most datasets come from the openly accessible UCI repository \cite{Dua2017}. 
We included five classification tasks and five regression tasks with different sample sizes and numbers of features. 
All classification problems were binary or were binarized, and class labels were balanced to exclude biases due to class imbalance \cite{Collell2018}.

We also used one clinical dataset, a balanced subsample of the ADHD speech dataset described by von Polier et al. \cite{Polier2021} includes 126 individuals with 6016 speech-related features, the binary target describing ADHD status (ADHD or control) and contains four confounds: gender, education level, age and, depression score measured using the Beck's depression inventory (BDI).
For more information on the datasets see Supplementary Table S1. 

\subsection{Confound removal }\label{sec4.2}
Confound removal was performed following the standard way of using linear regression models.
Following the common practice, we applied CR to all the features.
Specifically, for each feature, a linear regression model was fit with the feature as the dependent variable and the confounds as independent variables.
The residuals of these models, i.e., original feature minus the fitted values were used as confound-free features ($X_{CR}=X - \hat{X}$).
This procedure was performed in a CV-consistent fashion, i.e., the confound removal models were fitted on the training folds and applied to the training and test folds \cite{Snoek2019, More2021}.

\subsection{Machine Learning Pipeline }\label{sec4.3}
To study the effect of CR on both linear and nonlinear ML algorithms, we employed a variety of algorithms: linear/logistic regression (LR), linear kernel Support-vector machine (linear SVM), Radial Basis Function kernel Support-vector machine (RBF SVM), decision tree (DT), random forest (RF), and multilayer perceptron (MLP) with a single hidden layer (relu). Additionally, we used dummy models to evaluate chance-level performance.

In the preprocessing steps, we normalized the continuous features and continuous confounds to have a mean of zero and unit variance, again in a CV-consistent fashion.
Any categorical features were one-hot encoded following standard practice. 

\subsection{Evaluation}\label{sec4.4}
We compared the performance of ML pipelines with and without CR.
To this end, we computed the out-of-sample Area under the Curve for ROC (AUCROC) for classification and predictive $R^2$ from scikit-learn \cite{Buitinck2013} for regression problems in a 10 times repeated 5-fold nested CV. 
We employed the Bayesian ROPE approach \cite{Benavoli2017} to determine whether the results for a given dataset and algorithm with and without CR were meaningfully higher, lower or not meaningfully different.

\subsection{Predictability of $\hat{X}$}\label{sec4.5}
Whenever CR lead to an increase in performance this can only have one of two reasons: 
either 1) revealing information present in the features, or 
2) leaking confounding information. 
To reveal information in the features the CR  has to suppress variance in the features which make learning generalizable features-target relationship harder. 
For example, unrelated noise or linear shortcuts could be suppressed.
In other words, suppression works by removing less predictable variance in the data. 
This means that $\hat{X}$ has to be less predictive of the target than $X$ in the resulting CR-ML workflow.
If one finds contrasting evidence, an especially highly predictive  $\hat{X}$, this is strong direct evidence for confound-leakage through CR.

\subsection{Code}\label{sec4.7}
Additional information and code can be found under \url{https://github.com/juaml/ConfoundLeakage}.

\subsection*{Acknowledgement}\label{sec4.8}
This work was partly supported by the Helmholtz-AI project DeGen (ZT-I-PF-5-078), and the Helmholtz Portfolio Theme ‘Supercomputing and Modeling for the Human Brain’.
We thank the UCI  machine learning repository \cite{Dua2017} and the original dataset contributors.
Georg G. von Polier participated and received payments in the national advisory board ADHD of Takeda.

\bibliographystyle{unsrt}  
\bibliography{Reference}

\begin{FPfigure}
    \centering
    \includegraphics[width=.9\textwidth]{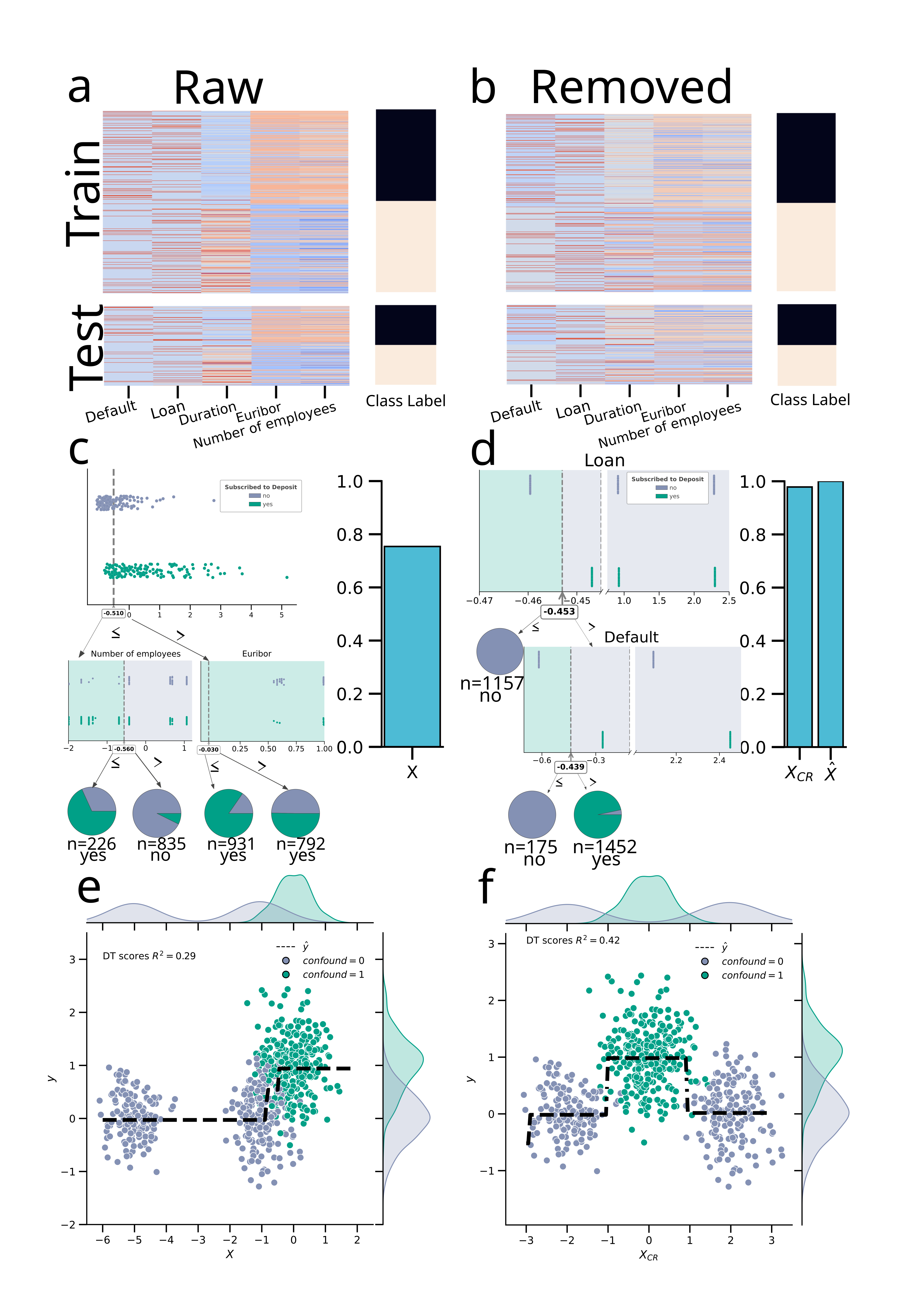}
    \caption{
A walk-through analysis demonstrating our analysis pipeline and confound-leakage using DT.
The results shown here are on the $30\%$ test split.
For the binary classification walk-through using the bank investment dataset, a subset of the features used are shown before CR (a) and after CR (b).
Induced DTs and their performance before (c) or after CR (d).
The DT after CR (d) is based on minute differences in only two features and still performs nearly perfectly and better compared to the DT on raw data (c).
The regression analysis walk-through using simulated data is depicted as feature-target relationships with the dotted line showing the predicted values (e,f).
The non-normal distribution of the feature conditioned on the confound leaks information usable by the DT.
Here, CR removes the linear relationship, as intended, but introduces a stronger non-linear one by shifting the distribution of $X_CR$ given $confound=0$ in-between the two peaks of $X_CR$ given $\text{confound}=1$ (f).
}
    \label{fig:walk}
\end{FPfigure}

\begin{FPfigure}
    \centering
    \includegraphics[width=.9\textwidth]{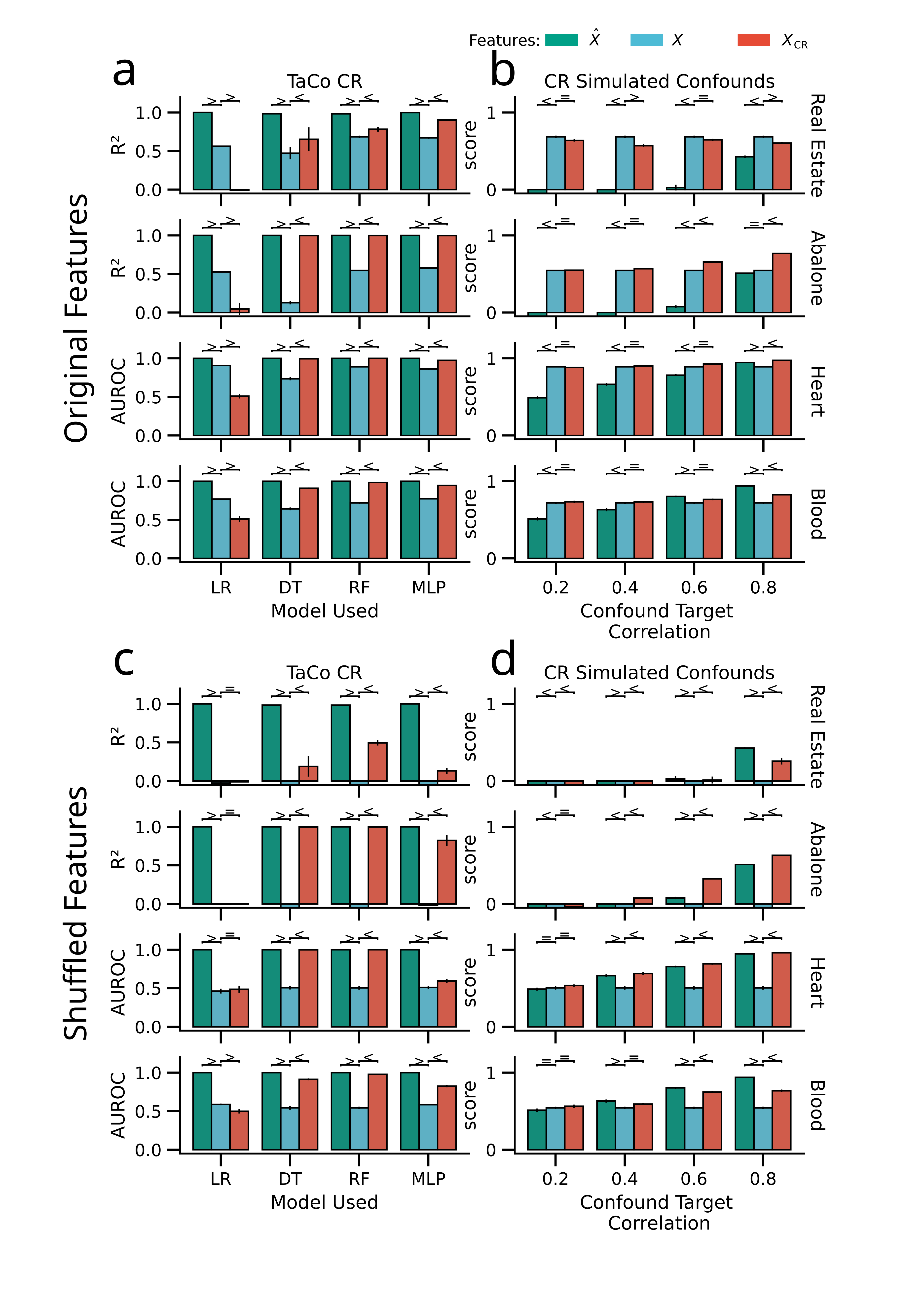}
    \caption{
Performance on the UCI benchmark datasets when using raw vs CR features (a) and raw vs the predicted features given the confound/TaCo/$\hat{X}$ (b).
The two columns correspond to: 
1) TaCo removal with four ML algorithms (LR, DT, RF, MLP), and  
2) CR with simulated confound with different correlation to the target (range 0.2-0.8) with RF.
(a,b) show performance using the original features while (c,d) show the performance on shuffled features.
When using a linear model (LR) TaCo removal leads to reduction in prediction performance, as expected. 
In contrast, nonlinear models lead to a higher performance for all datasets. 
This increase could be either explained by confound removal revealing information already in the data (suppression) or confound removal leaking information into the features (confound-leakage).
Shuffling the features destroys association between features and the target, therefore subsequent performance increase after TaCo removal indicates the possibility of confound-leakage (c,d). 
Additionally, the higher performance of $\hat{X}$ (a,b) compared to $X$ does not support suppression as explanation as suppression assumes that confound-removal removes noise or other at most weakly predictive variance from the features. 
In this case, the variance removed feature $\hat{X}$ should be less predictive than the raw features $X$. 
The simulated confounds show that an increase after CR is also possible for confounds weakly related to the target (b,d) and one dataset (Blood) shows strong evidence of confound-leakage.}
    \label{fig:uci}
\end{FPfigure}

\begin{FPfigure}
    \centering
    \includegraphics[width=.9\textwidth]{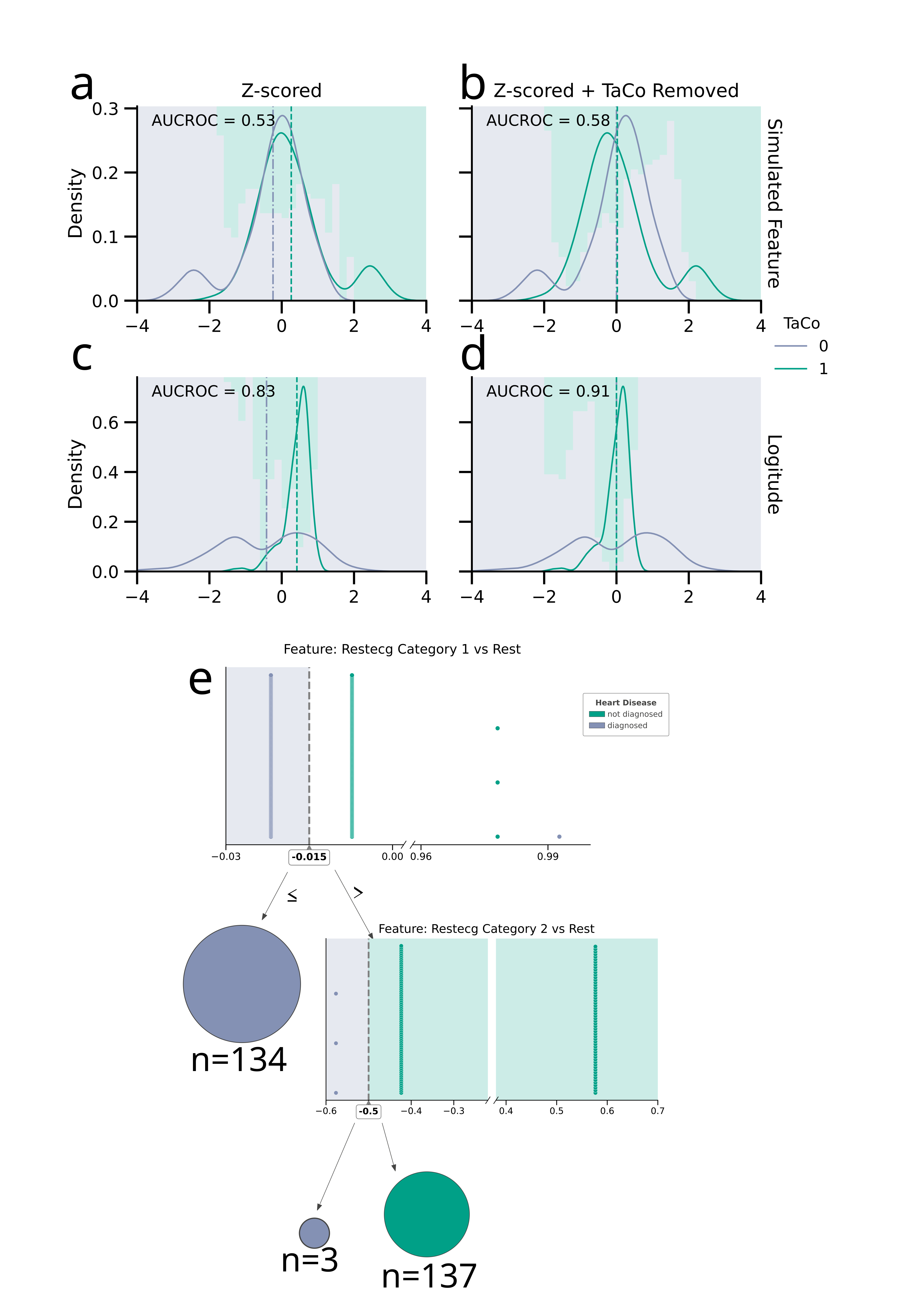}
    \caption{Two mechanisms for confound-leakage.
First mechanism where non-normal distributions get shifted apart through CR. (a,b) show this using a simulation with extreme values on opposing sides for one feature conditioned on the TaCo. 
(c,d) show a simplified version (binary target for visualization purposes) of the house price UCI benchmark dataset.
Here, the distributions of the feature conditional on the TaCo are different (c); a narrow distribution ($\text{TaCo}=1$) and a distribution with two peaks ($\text{TaCo}=0$).
TaCo removal shifts the narrow distribution in-between the two peaks (d), leaking information usable by non-linear ML algorithms. 
The second mechanism, leakage through minute differences in the feature after CR, is highlighted through the visualization of the DT trained on the heart dataset after CR (e). 
Distribution plots visualize the data at each decision node. 
The decision boundary is shown as a dotted line. 
For decision nodes before leaf nodes, the side of the decision node leading into a prediction is colored to represent the predicted label as diagnosed (green) or not (purple). 

The minute differences in the two used features that perfectly separate the data into the two classes can be seen.}
    \label{fig:sim}
\end{FPfigure}

\begin{FPfigure}
    \centering
    \includegraphics[width=.9\textwidth]{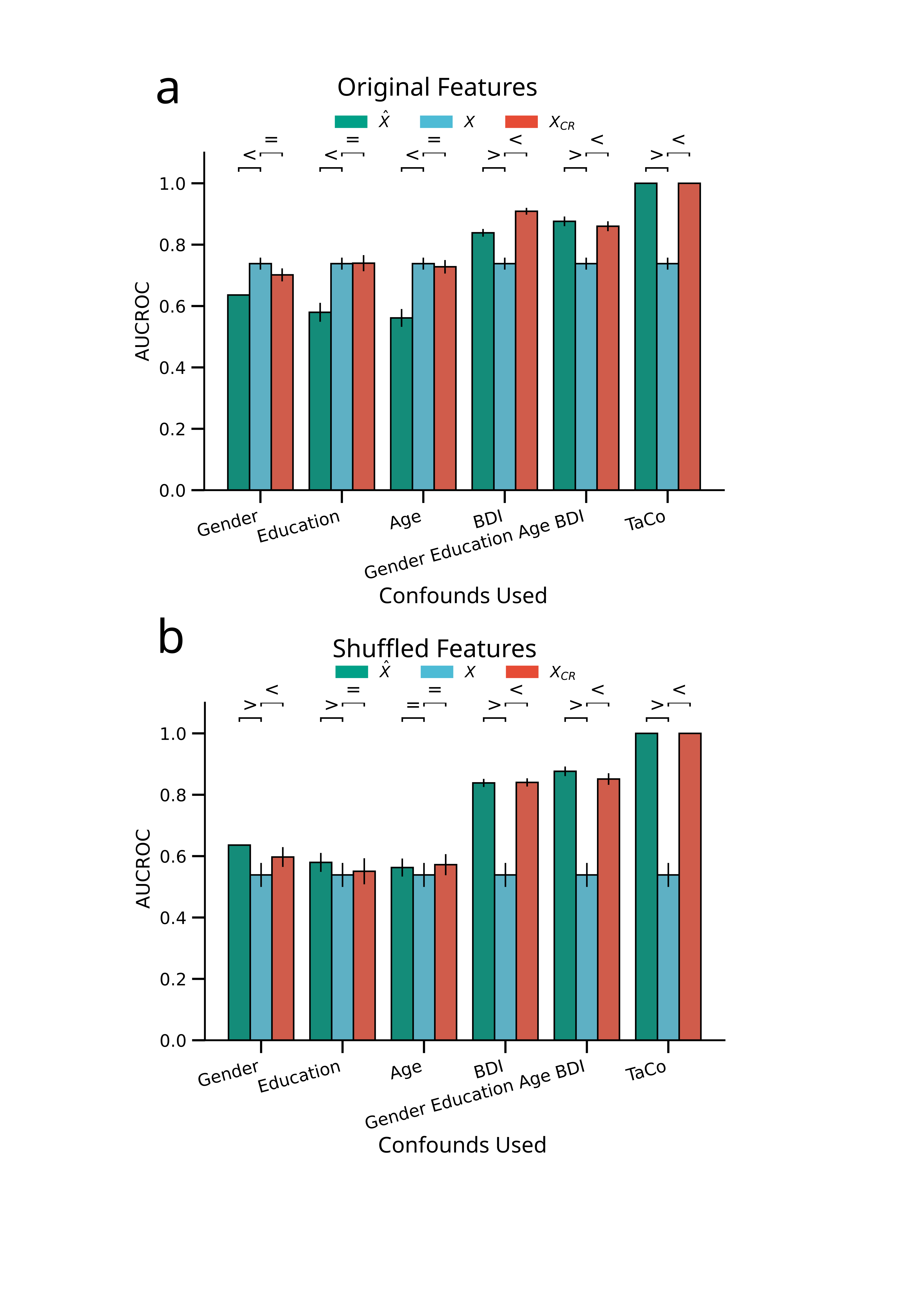}
    \caption{
Summary of the performance on the real-world ADHD speech dataset when using different confounds.
Note that the features used were always the same.
Increased performance, for both original and shuffled features, can be seen when using the TaCo and when BDI was used as a confound.
This suggests that BDI is driving the performance increase.}
    \label{fig:real}
\end{FPfigure}

\end{document}